\documentclass[10pt,twocolumn,letterpaper]{article}

\usepackage{times}
\usepackage{graphicx}
\usepackage{amsmath,amssymb,amsfonts}
\usepackage{booktabs}
\usepackage{caption}
\usepackage{subcaption}
\usepackage{xcolor}
\usepackage{url}
\usepackage{nicefrac}
\usepackage{microtype}
\usepackage{bbold}
\usepackage{comment}
\usepackage{lineno}

\definecolor{linkblue}{rgb}{0.21,0.49,0.74}
\usepackage[pagebackref,breaklinks,colorlinks,allcolors=linkblue]{hyperref}

\newcommand{\CelebCount}{48}
\newcommand{\CelebTrainFrames}{101{,}870}
\newcommand{\CelebTestFrames}{48{,}500}
\newcommand{\CelebVideos}{828}

\newcommand{\LPCount}{360}
\newcommand{\LPTrainFrames}{47{,}806}
\newcommand{\LPTestFrames}{18{,}292}
\newcommand{\LPVideos}{375}

\title{	
FANVID: A Benchmark for Face and License Plate Recognition in Low-Resolution Videos}

\author{%
\begin{tabular}{cccc}
\textbf{Kavitha Viswanathan} &
\textbf{Sanket Potdar} &
\textbf{Vrinda Goel} &
\textbf{Shlesh Gholap} \\
Dept. of Electrical Engineering & Dept. of Electrical Engineering & Dept. of Electrical Engineering & Dept. of Electrical Engineering \\
IIT Bombay, India & IIT Bombay, India & IIT Bombay, India & IIT Bombay, India \\
\texttt{184070024@iitb.ac.in} & \texttt{sanketpotdar.iitb@gmail.com} & \texttt{20d070090@iitb.ac.in} & \texttt{shleshgholap@iitb.ac.in} \\
\\
\textbf{Devayan Ghosh} &
\textbf{Madhav Gupta} &
\textbf{Dhruvi Ganatra} &
\textbf{Amit Sethi} \\
Dept. of Electrical Engineering & Dept. of Electrical Engineering & Dept. of Electrical Engineering & Dept. of Electrical Engineering \\
IIT Bombay, India & IIT Bombay, India & IIT Bombay, India & IIT Bombay, India \\
\texttt{23m1079@iitb.ac.in} & \texttt{21D070043@iitb.ac.in} & \texttt{21d070027@iitb.ac.in} & \texttt{asethi@iitb.ac.in} \\
\end{tabular}
}

\begin{document}

\maketitle

\begin{abstract}

Real-world surveillance often renders faces and license plates unrecognizable in individual low-resolution (LR) frames, hindering reliable identification. To advance temporal recognition models, we present FANVID, a novel video-based benchmark comprising 1,203 LR clips ($180 \times 320$, 20–60 FPS) featuring 48 identities and 360 license plates. Each video includes distractor faces and plates, increasing task difficulty and realism. The dataset contains over 216,000 manually verified bounding boxes and labels.

FANVID defines two tasks: (1) face matching—detecting LR faces and matching them to high-resolution mugshots, and (2) license plate recognition—extracting text from LR plates without a predefined database. Videos are downsampled from high-resolution sources to ensure that faces and text are indecipherable in single frames, requiring models to exploit temporal information. We introduce evaluation metrics adapted from mean Average Precision at IoU > 0.5 (mAP@0.5), prioritizing identity correctness for faces and character-level accuracy for text.

Baseline method with pre-trained video super-resolution, detection, and recognition achieved performance scores of 0.58 (face matching) and 0.42 (plate recognition), highlighting both the feasibility and challenge of the tasks. FANVID's selection of faces and plates balances diversity with recognition challenge. We release the software for data access, evaluation, baseline, and annotation to support reproducibility and extension. FANVID aims to catalyze innovation in temporal modeling for LR recognition, with applications in surveillance, forensics, and autonomous vehicles.

\end{abstract}

\section{Introduction}

Surveillance cameras, now ubiquitous in urban environments, often produce low-resolution (LR) footage where faces and license plates are unrecognizable in single frames. A March 2025 BBC report noted that many images from CCTV cameras are too grainy for reliable facial recognition~\cite{bbc2025face}. While judicial evidence must rely on high resolution (HR), there is an urgent need for robust LR recognition methods that can narrow the search for vehicles, suspects, and missing persons.

Most of the existing recognition models and benchmarks such as SCFace~\cite{grgic2011scface} rely on static HR images, neglecting the potential of temporal context in LR sequences. To address this gap, we introduce FANVID (Face And Number identification in low-resolution VIDeos), a new benchmark for recognizing faces and license plates in LR videos ($320\times180$, 20–60 FPS), where individual frames are indecipherable to the naked eye.

FANVID includes 1,203 videos spanning 48 identities and 360 license plates, totaling more than 216,000 manually verified bounding boxes and labels. The benchmark defines two tasks: (1) Face Matching: detecting LR faces in video clips and matching them to HR mugshots; and (2) License Plate Recognition: detecting and transcribing LR plate text without a prior database. Each video includes distractor faces and plates to simulate real-world conditions. The FANVID dataset is hosted on \href{https://huggingface.co}{HuggingFace}.

We propose task-specific evaluation metrics, adapted from mean Average Precision at IoU > 0.5 (mAP@0.5), with identity-level correctness for face matching and character-level accuracy (via normalized edit distance) for plates. FANVID videos are downsampled from HR sources, and all annotations were manually verified pre-downsampling to ensure high-quality labels. Baseline pipelines using RCDM~\cite{viswanathan2024rcdm} for video super-resolution, followed by RetinaFace~\cite{deng2020retinaface} and InsightFace~\cite{deng2019InsightFace} for face, and EasyOCR~\cite{easyocr2020} for text achieve accuracy scores of 0.58 and 0.42, respectively, underscoring the challenge and opportunity for temporal modeling approaches, with or without video super-resolution.

FANVID balances demographic and textual diversity across races, genders, cultures, and license plate formats with recognition confusion. To promote responsible research, it uses links to public-domain videos and includes tools for dataset creation and annotation. Our contributions include:
\begin{enumerate}
    \item A curated benchmark of over 216,000 annotations and identity/text labels across 1,203 videos.
    \item Task-adapted metrics and baseline using RCDM, RetinaFace, InsightFace, and EasyOCR.
    \item Evaluation scripts and data generation tools (using SAM2) for reproducibility and extension.
\end{enumerate}

FANVID is designed primarily for evaluation but includes open-source tooling to support future dataset growth for additional challenges and extended training. Its size supports both lightweight and deep models, and its focus on realistic data makes it ideal for advancing research in surveillance, forensics, and autonomous driving. We believe FANVID will catalyze progress in temporal modeling for LR recognition while fostering inclusive, reproducible, and application-driven research.

\section{Related Work}

Several datasets have advanced the state-of-the-art in face and license plate detection, recognition, and verification, as well as super-resolution. However, existing datasets either lack temporal information, realistic low-resolution conditions, distractors (other faces and plates), or task structures that reflect real-world surveillance and forensics needs. 

\subsection{Face and License Plate Recognition Datasets}

Face and license plate recognition have been extensively studied in computer vision. Datasets like LFW~\cite{huang2008labeled}, MegaFace~\cite{kemelmacher2016megaface}, and IJB-C~\cite{maze2018iarpa} target unconstrained face recognition but rely on high-resolution (HR) still images or video frames with clear frontal views. Surveillance-oriented datasets such as SCFace~\cite{grgic2011scface} and QMUL-SurvFace~\cite{cheng2018surveillance} introduce LR face images from surveillance-style cameras but are limited to static frames and lack temporal information that could aid recognition. Similarly, datasets like UFPR-ALPR~\cite{silva2018license}, OpenALPR, and the VIVA Challenge~\cite{jodoin2016tracking} provide real-world license plate images, but these are often HR or only mildly degraded, and generally do not include sequences that support temporal modeling.


\subsection{Recognition in Video: Face, Person, and Text}

Video-based recognition offers an opportunity to overcome the limitations of LR in single frames by aggregating information across time. Benchmarks such as Celebrity-1000~\cite{liu2018dependency}, IJB-S~\cite{maze2018iarpa}, and MARS~\cite{zheng2016mars} facilitate video-based face or person recognition, but their footage is typically of higher quality. These datasets do not present the joint challenge of cross-resolution, cross modality (video and image), and cross-scene (different cameras, clothing, backdrop) identity matching.

For license plates, most benchmarks focus on character detection in isolated images or frames. Recognition of text in videos has been explored in datasets like COCO-Text~\cite{veit2016coco} and ICDAR video challenges~\cite{karatzas2015icdar}, but these focus on scene text rather than license plates, and they assume sufficient resolution per frame for accurate OCR. 
The few datasets that support text recognition over frames (e.g., TextVQA~\cite{singh2019towards}) feature high-quality imagery. 

\subsection{Video Super-Resolution Datasets and Their Applicability}

Video super-resolution (VSR) datasets like Vid4~\cite{liu2011bayesian}, Vimeo-90K~\cite{xue2019video}, and REDS~\cite{nah2019ntire} have driven progress in frame enhancement, but their main task is to enhance perceptual quality. 
Consequently, they are poorly suited for evaluating how well VSR benefits downstream recognition.
While joint SR-recognition pipelines have shown some promise~\cite{wang2021real}, their effectiveness diminishes when frame content is severely degraded or when objects appear small or distorted in real surveillance footage.

FANVID bridges these gaps and pioneers a new class of benchmarks for identity verification and text recognition from temporal information in realistic low-resolution and distractor-rich videos.

\subsection{Related Datasets and Benchmarks (Extended)}
\label{sec:related-extended}
\vspace{2pt}
\noindent\textbf{Faces (images→video).}
LFW, CelebA, VGGFace2, MS-Celeb-1M catalyzed unconstrained face recognition but skew to still, relatively clean imagery. YouTube Faces and IJB-S add temporal variation yet assume medium–high resolution per frame. Surveillance-oriented SCFace/QMUL-SurvFace introduce LR but under static conditions or with sparse temporal labels.

\noindent\textbf{License plates (images).}
UFPR-ALPR, AOLP, SSIG-SegPlate benchmark detection/OCR under traffic scenes; many assume near-frontal plates, limited distractors, and Latin scripts.

\noindent\textbf{Systemic gaps addressed by FANVID.}
(i) \emph{Temporal realism at LR} where no single frame suffices,
(ii) \emph{Open-set distractors} for both modalities,
(iii) \emph{Task-aligned metrics} (localization+ID; normalized character accuracy),
(iv) \emph{Reproducibility} (links+scripts+schemas), and
(v) \emph{Diversity} targets (identities/plates; future non-Latin scripts).

\begin{table*}[t]
\centering
\caption{Positioning FANVID among representative datasets. “Temporal LR” = recognition requires aggregating across LR frames; “Distractors” = non-target entities present; “Reproducible pipeline” = links, schemas, and code to regenerate LR/annotations.}
\label{tab:relwork-compare}
\resizebox{\textwidth}{!}{
\begin{tabular}{lccccc}
\toprule
\textbf{Dataset} & \textbf{Modality} & \textbf{Still vs Video} & \textbf{Temporal LR} & \textbf{Distractors} & \textbf{Reproducible pipeline} \\
\midrule
LFW / VGGFace2 / MS-Celeb-1M & Face & Still & \(\times\) & \(\triangle\) & \(\triangle\) \\
YouTube Faces / IJB-S & Face & Video & \(\triangle\) (not severe) & \(\triangle\) & \(\triangle\) \\
SCFace / QMUL-SurvFace & Face & Images/Tracks & \(\triangle\) (static/partial) & \(\triangle\) & \(\triangle\) \\
UFPR-ALPR / AOLP / SSIG-SegPlate & LP & Images & \(\times\) & \(\triangle\) & \(\triangle\) \\
\textbf{FANVID (ours)} & Face+LP & \textbf{Video} & \textbf{\checkmark} & \textbf{\checkmark} & \textbf{\checkmark} \\
\bottomrule
\end{tabular}}
\end{table*}

\section{FANVID Dataset}

FANVID (Face And Number identification in low-resolution VIDeos) is a new benchmark dataset to support two real-world tasks: (1) face matching, which involves identifying individuals from LR video against HR mugshots, and (2) license plate recognition, which requires transcribing plate text without external databases. 
It emphasizes using temporal cues in realistic, challenging conditions where individual frames are often indecipherable and distractors are present. It contains over 1,400 video clips of faces and license plates sourced from public HR footage for \emph{manual labeling} and downsampled to mimic surveillance-like quality. 
In the following subsections, we describe the dataset creation and annotation process and its characteristics, and the tools provided for reproducibility and extension.

\subsection{Dataset Creation}

FANVID was constructed using high-resolution (HR) videos sourced from YouTube\textsuperscript{\textregistered}, specifically focusing on content resembling surveillance footage—such as citizen-recorded street scenes, security camera footage, or candid public appearances of celebrities. All source videos were restricted to publicly available content to ensure traceability and compliance with platform guidelines. To establish identity ground truth, mugshots or frontal HR images of each subject (e.g., celebrities) were obtained from different scenes, such as press events or interviews.

\paragraph{Face Annotation} Facial bounding boxes were initially generated using RetinaFace~\cite{deng2020retinaface} detector and InsightFace~\cite{deng2019InsightFace} feature embedding was used for identity matching with the mugshot gallery. This automated step produced provisional bounding boxes and subject identity labels across video frames. These annotations were then manually verified and corrected \emph{frame-by-frame} to ensure high-precision localization and identity tagging, especially under occlusion, extreme pose, or low visibility conditions. Non-target faces (distractors) appearing in the scenes were not annotated.

\paragraph{License Plate Annotation.} For license plates, a single high-quality frame was manually annotated with a bounding box around the plate region. This annotation was then propagated across subsequent frames using SAM2~\cite{sam2_2024}.
Bounded boxes were manually examined and corrected, if needed. Plate text was transcribed manually from the clearest HR frames, ensuring that characters were correct even in cases of partial occlusion or motion blur.

\paragraph{Low-Resolution Conversion.} After verification and corrections, all videos were uniformly downsampled using OpenCV (bicubic) to a spatial resolution of $320 \times 180$. The clip selection and downsampling parameters were calibrated to ensure that neither facial identity nor license plate text is human-recognizable in \emph{any individual frame}, enforcing reliance on temporal context for recognition. 

\subsection{Dataset Description}

FANVID comprises over 31,096 annotated frames spanning 1,463 videos. It represents 63 persons selected to balance racial (including South Asian, East Asian, Black, White, Mori, Latino, and mixed) and gender diversity with appearance similarity. Each person can appear in multiple clips, sometimes under different conditions or camera angles, to test generalization.

For license plate side, the dataset includes 49 unique plates annotated across several countries, with variations in structure, color, and visibility. Although we restricted to English, plates vary by color (white or yellow background), length (five to nine characters), and content (alphanumeric versus numeric). Plates may be fragmented or occluded in multiple frames.

The dataset has been split into predefined \texttt{train} (including validation), and \texttt{test} sets. No individual or plate appears in more than one split to enforce generalization. Splits are balance by difficulty and identity classes or plate characteristics.

\begin{figure*}[htbp]
    \centering


    \begin{subfigure}[t]{0.48\textwidth}
        \includegraphics[trim=50 500 260 80, clip, width=.9\linewidth]{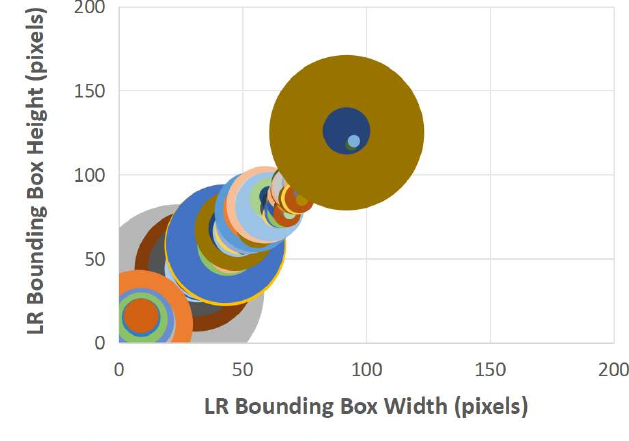}
        \caption{Bounding box dimensions for FANVID faces}
        \label{fig:bbox_dimensions_face}
    \end{subfigure}
    \hfill
    \begin{subfigure}[t]{0.48\textwidth}
        \includegraphics[trim=50 500 220 80, clip,width=\linewidth]{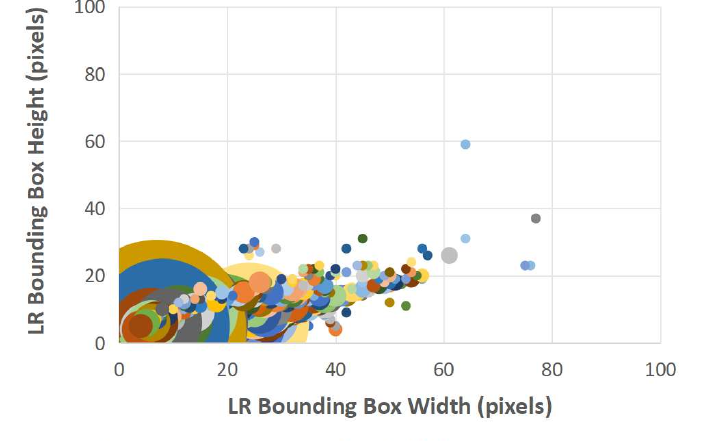}
        \caption{Bounding box dimensions for FANVID plates}
        \label{fig:bbox_dimensions}
    \end{subfigure}

    \caption{Bounding box dimension distribution in FANVID all LR frames (frequency is shown by bubble area) indicate that most faces and plates are too small for human recognition.}
    \label{fig:fanvid_all_plots}
\end{figure*}

\begin{figure*}[t]
    \centering
    \includegraphics[trim=80 580 120 10, clip, width=0.9\textwidth]{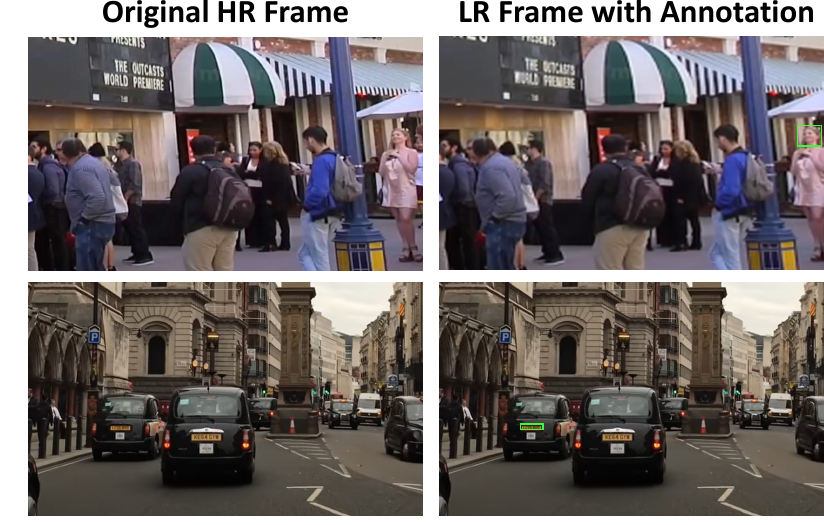}
    \caption{Example frames with HR reference, downscaled LR (rescaled for clarity), and annotated bounding boxes for faces and license plates.}
    \label{fig:example_images}
\end{figure*}

Table~\ref{tab:DataChar} summarizes data characteristics and Figure~\ref{fig:fanvid_all_plots} show distribution of LR bounding boxes. Additional data characteristics are given in Supplementary Materials. The current version for faces is FANVID-Faces-1 and that for text is FANVID-Text-1, with plans for releasing larger and even more challenging future versions with community participation. Figure~\ref{fig:example_images} shows a few examples of original HR frames, their LR versions along with annotated bounding boxes.

\begin{table}[t]
\centering
\caption{Summary of FANVID Dataset Characteristics}
\label{tab:DataChar}
\resizebox{\linewidth}{!}{%
\begin{tabular}{lcc l}
\toprule
\textbf{Characteristic} & \textbf{FANVID-Faces-1} & \textbf{FANVID-Text-1} & \textbf{Notes} \\
\midrule
Frames (train) & \CelebTrainFrames & \LPTrainFrames & LR annotated frames (train) \\
Frames (test)  & \CelebTestFrames  & \LPTestFrames  & Held-out, video-level split \\
Video clips    & \CelebVideos      & \LPVideos      & Diverse backgrounds \\
Entities (IDs) & \CelebCount       & \LPCount       & Unique faces / plates \\
LR frame size  & $320\times180$    & $320\times180$ & Fixed size \\
Normalized Gini coeff. & 0.192 (race), 0.308 (gender) & 0.436 (country) & Plates from US dominate \\
\bottomrule
\end{tabular}%
}
\end{table}

\subsection{Data and Code Availability}

To promote tansparency and responsible data use, FANVID is distributed as a separate CSV file for each task. Each row in these files corresponds to a bounding box with the following fields:

\begin{itemize}
    \item \textbf{Video URL}: To access the original high-resolution YouTube\textsuperscript{\textregistered} video.
    \item \textbf{Metadata:} Resolution (typically, 1020p or 720p) and frame rate for HR source video.
    \item \textbf{Frame Number}: Index of the frame in the original video applicable to the bounding box.
    \item \textbf{Crop Coordinates (optional)}: Coordinates specifying any cropping needed, if a simple downscaling is insufficient to make the faces or text unrecognizable.
    \item \textbf{Low-Resolution Frame Size}: Traget LR fixed at $320 \times 180$ for all clips.
    \item \textbf{Bounding Box Coordinates}: Location of the annotated entity within the LR frame.
    \item \textbf{Entity Identifier}: Name of the individual (for faces) or the transcribed license plate text.
    \item  \textbf{Split:} To indicate if an entity should be used for training (including validation) or testing.
\end{itemize}

To enable face matching across scenes, we also provide an additional file with rows corresponding to each person in the video dataset with a link to a single publicly available high-resolution mugshot obtained from scenes not overlapping the videos.

To respect the copyright of the original video owners, FANVID does not redistribute the video files. Instead, for reproducibility, we provide code for frame extraction, cropping, downscaling, and annotation based on the provided CSV files. 
FANVID also follows the “Datasheets for Datasets” best practice for documentation, outlining collection methods, licensing, intended use, and ethical considerations.

In addition to the test set used for benchmarking, we also provide a carefully curated training set for both modalities. This allows researchers to explore multiple avenues: (i) developing and training novel architectures directly on FANVID to test their ability to generalize under realistic low-resolution conditions, and (ii) applying standard off-the-shelf methods trained on other corpora to evaluate zero-shot transfer performance. By including both train and test splits, our benchmark not only supports rigorous evaluation but also encourages innovation in model design, domain adaptation, and temporal representation learning. Researchers can therefore choose to either validate their approaches against established baselines or leverage the training data to push the boundary of low-resolution recognition.



We also release tools for extending the dataset for additional training, such as by adding new identities, text, or sequences. This includes full codebase used for the automated portion of the annotation pipeline, including VSR, face detection and matching, and plate detection and tracking, and structured annotation generation. However, manual verification remains essential for ensuring annotation quality.

We recommend that practitioners avoid splitting data (videos, mugshots) of the same individual or plate into training and testing. This avoids identity-specific overfitting in recognition results. We also recommend running each experiment multiple times, such as with different random weight initializations, to report standard error.

\section{Tasks, Metrics, and Baseline Results}

FANVID defines two challenging recognition tasks that require recognition over low-resolution (LR) video sequences in the presence of distractors and limited per-frame information. Unlike previous benchmarks that focus on frame-level classification or HR imagery, these tasks demand temporal aggregation and robust cross-domain matching. In this section, we describe the recognition tasks, outline our evaluation metrics, and present baseline results.

\subsection{FANVID Tasks}

FANVID is designed for two real-world surveillance and forensic recognition tasks:

\paragraph{1. Face Matching:} Given a set of HR reference images (e.g., mugshots) and a query LR video containing one or more faces, the goal is to detect and identify the target individual in the video. Each video may include multiple distractor faces not present in the gallery, and individual frames are too low-resolution for reliable recognition without temporal aggregation. \emph{Each mugshot in the gallery must be matched to all locations and all clips to simulate a realistic search for a missing person across scenes.} The task therefore tests the ability of models to perform many-to-one face matching under cross-resolution and cross-domain conditions.

\paragraph{2. License Plate Recognition:} In this task, the goal is to detect the license plate region in each frame and transcribe the plate text from LR video without using an external database. Plates may be partially occluded or motion-blurred in many frames, making character-level temporal reasoning necessary. Plates must be detected without using their ground truth number for matching. Therefore, false positives are not penalized. Unlike conventional OCR benchmarks, this task does not assume access to high-quality stills or full plate visibility at all times.


\subsection{Evaluation Metrics}

We propose one metric each for face and text recognition. The metrics derived from mean average precision at IoU>0.5 (mAP@0.5), which is appropriately lenient on reasonable deviations in bounding boxes but penalizes recognition errors. While recognition tasks assume a fixed number of classes, our metric for face matching (also known as verification) needs to adapt to a variable number of reference mugshots in the database and a variable number of faces in an LR frame. Similarly, our text recognition metric must incorporate character-level accuracy normalized by plate text length.

\subsection{Scoring Protocol (Implementation Detail)}
\label{sec:scoring-detail}
We score \emph{per row} after grouping by \texttt{(VideoID, ClipID, FrameNo)}. Here \textbf{VideoID} is the celeb identity or plate string.

\paragraph{Celeb (face) rows.}
(1) Compute IoU between each detected bbox and the GT bbox at that frame; map to spatial credit
$b(\mathrm{IoU})=\min\!\left(1,\ \mathrm{IoU}/0.8\right)$ so $0.8\!\to\!1$.
(2) Extract a face embedding from the detected bbox and compare with the stored celeb embedding for \texttt{VideoID}; set $\mathbb{1}_{\mathrm{id}}\!=\!1$ iff identity matches; else $0.5$ if row score is $>0$.
(3) Row score: $\mathrm{FaceScore}=b(\mathrm{IoU})\cdot \mathbb{1}_{\mathrm{id}}$.
Save all row scores; report per-video means and dataset mean ($\pm$ s.e.).

\paragraph{License plate rows.}
(1) OCR the detected bbox to obtain $\hat{s}$.
(2) With ground-truth plate string $s$ (from \texttt{VideoID}), compute
$\mathrm{PlateScore} = 1 - \frac{\mathrm{EditDist}(\hat{s}, s)}{|s|}$,
after normalizing case and dropping non-alphanumerics for both.
Aggregate per-video and dataset-level as above.


With these motivations, we propose the following metric. Let ground truth boxes be indexed by $i \in \{1, ... , n\}$, and detected boxes be indexed by $j \in \{1, ... , m\}$. Let $I_{i,j} \in [0,1]$ be their $IoU$ and $F_{i,j} \in \{0,1\}$ be their face identity match indicator, and $T_{i,j} \in [0,1]$ be their text match score based on annotations. The difference between a binary face match indicator and a continuous text match score is to allow for minor text recognition errors in the absence of a database of license plate texts or images, unlike the access to a database of identity mugshots. We implemented $T_{i,j}$ as inverse normalized edit distance, i.e., one minus the ratio of edit distance (from detected to the ground truth text) to the length of the ground truth text (after capitalizing and removing only alphanumeric characters). For every ground truth box, we simply set the matching indicator with most overlapping detected box (with randomized tie breaker), provided the overlap is at least 0.5 (non-overlapping ground truth boxes, such as those in FANVID, ensure that each detected box can have at most one corresponding ground truth box with IoU>0.5), as $M_{i,j} = \mathbb{1}_{I_{i,j} \geq 0.5} \times \mathbb{1}_{j = argmax_{k}(I_{i,k})}$, where $\mathbb{1}_{condition}$ is indicator function that takes value 1 when the condition is true. This gives false negative indicators as $N_{i} = \mathbb{1}_{max_j(M_{i,j})=0}$, and false positive indicators as $P_{j} = \mathbb{1}_{max_i(M_{i,j})=0}$.

An evaluation script that compares boxes in a detection sheet with those in a ground truth sheet is provided to implement the proposed metrics in a consistent manner.

Thus, the proposed metric for the face detection and matching is:
\begin{equation}
\label{eq:face}
    FaceRecBox=\frac{\sum_{i,j} M_{i,j} F_{i,j}}{\sum_{i,j} M_{i,j} + \sum_{i} N_{i} + \sum_{j} P_{j}},
\end{equation}

The denominator metric for the proposed text detection and recognition metric excludes false positives counts due to the lack of access to possible plate numbers, as follows:
\begin{equation}
\label{eq:text}
    TextRecBox=\frac{\sum_{i,j} M_{i,j} T_{i,j}}{\sum_{i,j} M_{i,j} + \sum_{i} N_{i}}.
\end{equation}

\subsection{Reproducibility and Tools}
In line with NeurIPS dataset guidelines, we release:
\begin{itemize}
    \item A metadata-rich CSV file linking video sources, annotation records, and gallery references,
    \item Frame extraction and annotation verification scripts,
    \item LR degradation generation pipelines,
    \item Evaluation scripts compatible with the proposed IoU-tolerant metrics,
    
\end{itemize}

All components of FANVID are designed to be lightweight, extensible, and reproducible, and the entire inference baseline can be run on a single A100 80GB GPU in less than half a day. Researchers may augment the dataset using additional clips via the provided annotation code and degradation modules. The full dataset and tools will be released under a CC-BY-NC license upon publication.

\subsection{Baseline Recognition Pipelines and Results}

To establish strong and modular baselines, we designed a pipeline that combines models for video super-resolution, object detection,  and recognition — each selected to reflect state-of-the-art practices for resource-constrained deployment.

\begin{figure}[htbp]
    \centering
    \includegraphics[width=0.95\linewidth]{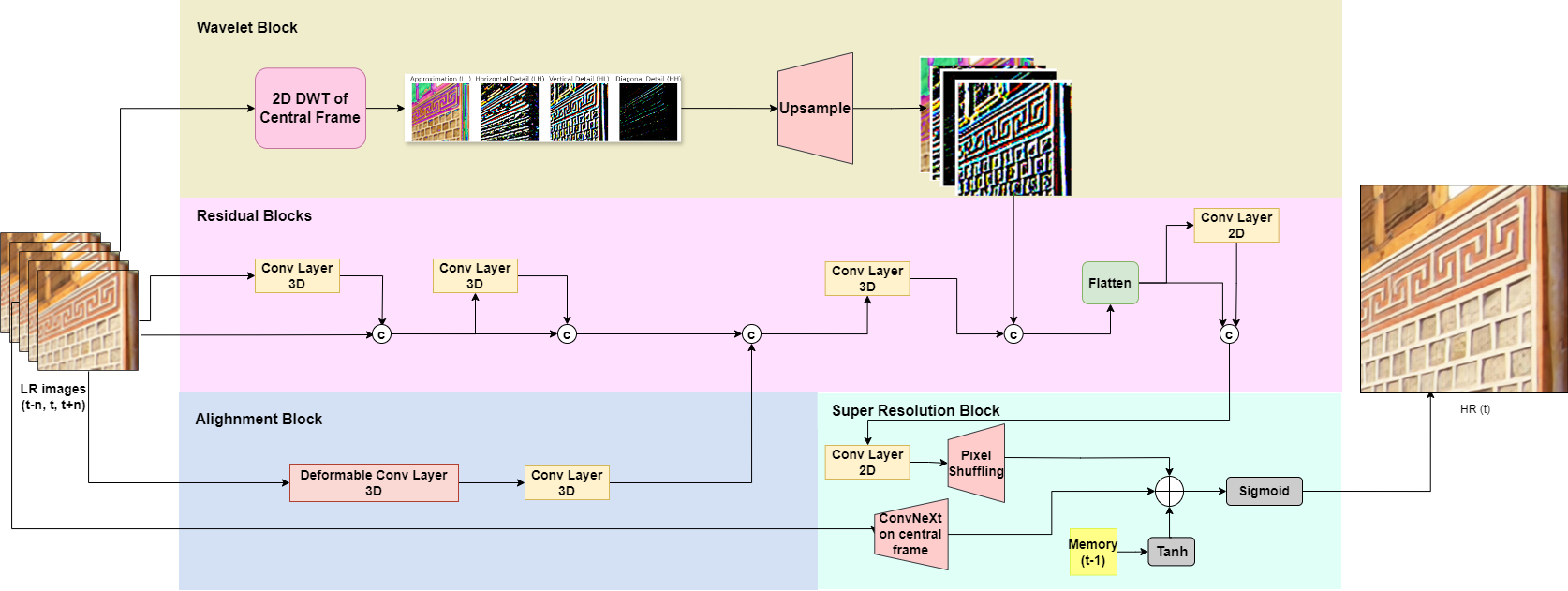}
    \caption{Architecture of the Residual ConvNeXt Deformable Convolution with Memory (RCDM) VSR model~\cite{viswanathan2024rcdm}, which integrates ConvNeXt blocks, deformable alignment modules, multi-level wavelet feature extraction, and temporal memory propagation for low-resource video enhancement (reproduced with permission).}
    \label{fig:rcdm_architecture}
\end{figure}

For all recognition tasks, low-resolution video input was first processed through RCDM \cite{viswanathan2024rcdm}, a memory-efficient video super-resolution model. RCDM employs deformable convolutions for inter-frame alignment, 2D wavelet decomposition for structural sparsity, and temporal memory units for consistent frame enhancement, as shown in Figure~\ref{fig:rcdm_architecture}. It produces a super-resolved frame sequence that serves as the input for downstream modules.

For face recognition, we used RetinaFace \cite{deng2020retinaface} for per-frame detection, followed by InsightFace \cite{deng2019InsightFace} for identity embedding. Feature vectors across frames were averaged, and cosine similarity was used to match the resulting video-level embedding to the mugshot gallery. This approach demonstrates robustness to motion blur, occlusion, and scale changes, with minimal temporal smoothing.

For license plate recognition, we used YOLOv10~\cite{wang2024yolov10} for bounding-box detection, leveraging its low-latency architecture and improved spatial precision. SAM2~\cite{sam2_2024} was used to track plate regions in subsequent frames to overcome blur, partial occlusion, and mismatch to distractor plates. The segmented regions were then passed to EasyOCR \cite{easyocr2020}, a multilingual text recognition engine, which extracts character-level transcription. Temporal smoothing of text transcriptions was done to obtain better results for plates tracked across multiple frames.

Modularity of the pipeline enables ablations across components and features (e.g. temporal fusion). Baseline code and settings are provided with the dataset to ensure reproducibility.



Results of baseline experiments are shown in Table~\ref{tab:results}
\begin{table}[t]
\centering
\caption{Baseline results on FANVID. Without video super-resolution (\textbf{RCDM}), performance drops significantly.}
\label{tab:results}
\resizebox{\linewidth}{!}{%
\begin{tabular}{lcccc}
\toprule
\textbf{Dataset} & \textbf{Split} & \textbf{Metric} & \textbf{Value} & \textbf{Technique} \\
\midrule
FANVID-Faces-1 & Test & FaceRecBox~\ref{eq:face} & 0.58 & Pre-trained \textbf{RCDM} + RetinaFace + InsightFace \\
FANVID-Text-1  & Test & TextRecBox~\ref{eq:text} & 0.42 & Pre-trained \textbf{RCDM} + EasyOCR \\
FANVID-Faces-1 & Test & FaceRecBox~\ref{eq:face} & 0.19 & Pre-trained RetinaFace + InsightFace \\
FANVID-Text-1  & Test & TextRecBox~\ref{eq:text} & 0.15 & Pre-trained EasyOCR \\
\bottomrule
\end{tabular}%
}
\end{table}

Beyond the baseline numbers in Table~\ref{tab:results}, our experiments reveal an interesting modality-specific trend. Using our video super-resolution (VSR) backbone (RCDM) in conjunction with standard recognition and detection pipelines substantially improves performance for the \emph{celebrity face matching task}, indicating that temporal enhancement can recover sufficient facial detail to boost identity discrimination. However, the \emph{license plate recognition task} does not benefit to the same extent. This is likely because our RCDM backbone, while lightweight and memory-efficient, is not powerful enough to recover fine-grained alphanumeric structures critical for OCR. We also experimented with other state-of-the-art VSR pipelines (e.g., BasicVSR++, RVRT), but observed no significant improvements in license plate transcription. This suggests that the generalization capacity of current models remains limited, particularly for structured text in severely degraded sequences, and highlights the need for new architectures that explicitly prioritize small-object and text-specific restoration.

As shown in the table above, while the tested methods provide a strong baseline even in case of low resolution and motion blur, there is a clear need for better methods to recognize persons and text in LR videos. Failure cases often involve severe occlusion or extreme lighting. Plate recognition errors correlate with frame-level character smearing and plate curvature. 

\subsection{Ablations and Error Analysis}
\label{sec:ablations}
\textbf{Effect of VSR.} Removing RCDM reduces FaceScore from 0.58 to 0.19 and PlateScore from 0.42 to 0.15, confirming the utility of spatial enhancement for tiny faces/plates.

\textbf{OCR robustness.} Plate errors correlate with character smearing and oblique angles; normalizing plate width via homography can improve edit distance on oblique subsets.

\textbf{Distractor impact.} Clips with $\geq2$ distractor faces increase identity confusion; PlateScore drops when multiple plates co-occur due to occasional region switches (mitigated by SAM2 tracking).

\subsection{Benchmark Gaps Addressed by FANVID}
\label{sec:benchmark-gaps}
\vspace{2pt}
\noindent\textbf{Faces (images→video).}
Datasets such as LFW, CelebA, VGGFace2, and MS-Celeb-1M have driven unconstrained face recognition but are dominated by high-quality still imagery. YouTube Faces and IJB-S add temporal variation, yet individual frames remain medium–high resolution. Surveillance-oriented datasets like SCFace and QMUL-SurvFace move toward LR scenarios but remain largely static or sparsely annotated temporally.

\noindent\textbf{License plates (images).}
Benchmarks such as UFPR-ALPR, AOLP, and SSIG-SegPlate evaluate detection and OCR in traffic scenes, but most assume near-frontal plates, limited distractors, and Latin-script contexts, leaving temporal dynamics underexplored.

\noindent\textbf{Unique contributions of FANVID.}
FANVID addresses several systemic gaps in prior datasets:  
(i) \emph{Temporal realism at LR}, enforcing recognition across sequences where no single frame suffices;  
(ii) \emph{Open-set distractors}, with both non-target faces and plates naturally included;  
(iii) \emph{Task-aligned metrics}, combining spatial localization with identity/text correctness;  
(iv) \emph{Reproducibility}, via links, scripts, and schemas to regenerate LR data and annotations; and  
(v) \emph{Diversity targets}, spanning identities and plates with plans for expansion beyond current domains.

\begin{table*}[t]
\centering
\caption{Positioning FANVID relative to representative datasets. “Temporal LR” = recognition requires aggregating across LR frames; “Distractors” = non-target entities present; “Reproducible pipeline” = links, schemas, and code to regenerate LR/annotations.}
\label{tab:relwork-compare}
\resizebox{\textwidth}{!}{
\begin{tabular}{lccccc}
\toprule
\textbf{Dataset} & \textbf{Modality} & \textbf{Still vs Video} & \textbf{Temporal LR} & \textbf{Distractors} & \textbf{Reproducible pipeline} \\
\midrule
LFW / VGGFace2 / MS-Celeb-1M & Face & Still & \(\times\) & \(\triangle\) & \(\triangle\) \\
YouTube Faces / IJB-S & Face & Video & \(\triangle\) (not severe) & \(\triangle\) & \(\triangle\) \\
SCFace / QMUL-SurvFace & Face & Images/Tracks & \(\triangle\) (static/partial) & \(\triangle\) & \(\triangle\) \\
UFPR-ALPR / AOLP / SSIG-SegPlate & LP & Images & \(\times\) & \(\triangle\) & \(\triangle\) \\
\textbf{FANVID (ours)} & Face+LP & \textbf{Video} & \textbf{\checkmark} & \textbf{\checkmark} & \textbf{\checkmark} \\
\bottomrule
\end{tabular}}
\end{table*}

\section{Limitations}
FANVID, while offering a novel dataset for recognition in low-resolution videos, has several limitations. The degradations applied are synthetically generated through bicubic downsampling, and do not capture additional challenges, such as compression artifacts, inclement weather, and graininess under low-lighting. The dataset is taken from English-speaking countries, with limited representation of global scripts and scene backgrounds. Although annotations are provided per frame, they do not include occlusion labels. HR references are curated mugshots from different domains, which may not reflect real-world query conditions of amateur or police photographs. Additionally, while we provide scripts instead of redistributing raw video, the original content remains publicly accessible at the owner's discretion, necessitating responsible use in line with licensing and privacy guidelines.

\section{Conclusion and Future Work}

FANVID introduces a new benchmark for low-resolution video-based recognition that reflects real-world surveillance conditions — where individual frames are grossly insufficient, and identity cues emerge only through temporal integration. In this regard, FANVID is unlike prior datasets that focus on static, high-resolution imagery without distraction from other faces or text. We define two tasks — face matching and license plate recognition — from LR videos with manually verified annotations and HR references. To ensure reproducibility and expansion, we release well-organized and documented code for data extraction, annotation, baseline methods (combining video super-resolution, detection, and recognition), and metric evaluation. We also provide spreadsheets that work with these codes with links to videos, their downscaling and cropping parameters, bounding box coordinates with entity labels, entity diversity class, and test split indicators. We provide encouraging baseline results that highlight the advantage of multi-frame recognition approaches. We encourage exploration of end-to-end temporal models which may not even require super-resolution as an intermediary step.

Several directions remain open. Unified architectures that jointly learn detection and identity from video (e.g., transformer-based spatiotemporal backbones) could improve robustness. The dataset's distractor-rich, partially occluded sequences invite research into occlusion-tolerant and few-shot learning methods. Additionally, videos with other realistic callenges -- such as, compression artifacts, inclement weather, low-light noise -- or native low-resolution can be collected, provided that the original entity can be reliably labeled. 

By grounding recognition in realistic, low-fidelity conditions and offering a reproducible, extensible benchmark, FANVID provides a foundation for advancing temporally-aware recognition systems that are both effective and responsible.

\section*{Broader Impact and Ethical Considerations}

FANVID was constructed entirely from publicly available videos and images, without redistributing any copyrighted material. Instead, we provide scripts to access these resources directly, in line with prevailing norms. All bounding boxes and labels were manually verified to ensure annotation quality.

Rather than minimizing visual overlap between identities, we intentionally include individuals with natural resemblance — similar skin tone, hairstyle, or age group — to reflect real-world visual ambiguity. This design encourages models to learn more robust, temporally grounded identity representations and not merely overfit to superficial visual cues. At the same time, we are committed to maintaining diversity across race, gender, and age. The dataset includes individuals from varied races and genders, and our curation pipeline ensures proportional representation across these attributes. 

All videos were downsampled to prevent recognition in single frames, mitigating privacy risks. We strongly discourage use of FANVID in actual surveillance or law enforcement contexts without formal review and adherence to privacy regulations. FANVID is intended purely for research into fair, interpretable, and privacy-conscious recognition under real-world visual constraints.




FANVID aspires to become a foundational tool for the development of socially-responsible recognition technologies. We support continued dialog with the AI ethics and research communities and will iterate on dataset, protocols, and documentation in response to evolving societal standards.

\bibliographystyle{plain}
\bibliography{references}

\end{document}